# Evaluation of the Performance of the Markov Blanket Bayesian Classifier Algorithm

Michael G. Madden

Department of Information Technology
National University of Ireland, Galway
Galway, Ireland
michael.madden@nuigalway.ie

**Abstract**

The Markov Blanket Bayesian Classifier is a recently-proposed algorithm for construction of probabilistic classifiers. This paper presents an empirical comparison of the MBBC algorithm with three other Bayesian classifiers: Naïve Bayes, Tree-Augmented Naïve Bayes and a general Bayesian network. All of these are implemented using the K2 framework of Cooper and Herskovits. The classifiers are compared in terms of their performance (using simple accuracy measures and ROC curves) and speed, on a range of standard benchmark data sets. It is concluded that MBBC is competitive in terms of speed and accuracy with the other algorithms considered.

## 1  Introduction

The Markov Blanket Bayesian Classifier (MBBC) algorithm has been recently proposed in a paper by this author [16] as an alternative to other structures based on Bayesian networks for classification tasks. In that paper, the algorithm was described some preliminary experiments using it were discussed. (Note that it was referred to as the *Partial Bayesian Network* algorithm in that paper.)

This paper presents an extended description of the MBBC algorithm (in Section 3), along with the results of a more comprehensive set of experiments (in Section 4). By way of providing context for this research, Bayesian networks and Bayesian classifiers are discussed in this section, and the K2 algorithm [7] for inductive learning of Bayesian networks is described in Section 2.

Several algorithms have been developed over the past decade for inductive learning of Bayesian networks. Russell and Norvig [22] identify four classes of problem, according to whether or not the structure is known and whether all variables are observed or some are hidden. In this paper we are concerned with problems where the structure is unknown and all variables are observed, and the resulting Bayesian network is applied to classification problems.

Although general Bayesian network structures may be used for classification tasks, as will be described in Section 2.3, this may be computationally inefficient since the classification node is not explicitly identified and not all of the structure may be relevant for classification, since parts of the structure outside of the classification node's Markov blanket (see Section 3). Accordingly, several simplified Bayesian structures, intended specifically for classification tasks, have been proposed; these include Naïve Bayes [14], Tree-Augmented Naïve Bayes [8] and Bayesian Network Augmented Naïve Bayes [4], which are illustrated in Figure 1 and discussed below in Section 1.2. However, in all of these structures it is assumed that the classification variable is the root node, thereby excluding structures where the classification variable is causally dependent on another variable. MBBC does not make that assumption. In addition, MBBC is more expressive than those other structures, as it is able to represent the full Markov blanket around a classification node in a general Bayesian network.

### 1.1  Bayesian Networks

Bayesian networks graphically represent the joint probability distribution of a set of random variables. A Bayesian network is composed of a qualitative portion (its structure) and a quantitative portion (its conditional probabilities). The structure $B_S$ is a directed acyclic graph where the nodes correspond to domain variables $x_1, \ldots, x_n$ and the arcs between nodes represent direct dependencies between the variables. Likewise, the absence of an arc between two nodes $x_1$ and $x_2$ represents that $x_2$ is independent of $x_1$ given its parents in $B_S$. Following the notation of Cooper and Herskovits [7], the set of parents of a node $x_i$ in $B_S$ is denote $\boldsymbol{p}_i$. The structure is annotated with a set of conditional probabilities ($B_P$), containing a term $P(x_i=X_i|\boldsymbol{p}_i=\boldsymbol{P}_i)$ for each possible value $X_i$ of $x_i$ and each possible instantiation $\boldsymbol{P}_i$ of $\boldsymbol{p}_i$.

Bayesian networks were originally constructed manually through consideration of causal dependencies in a system. However, several algorithms have been proposed in the last decade for inductive learning of Bayesian networks, following on from a seminal paper by Cooper and Herskovits [7]. Their K2 algorithm for Bayesian network induction is overviewed in Section 2, as the MBBC algorithm is based on it.

## 1.2 Classifiers based on Bayesian Networks

Figure 1 schematically illustrates the structure of the Bayesian classifiers considered in this paper. The simplest form of Bayesian classifier, known as *Naïve Bayes*, was shown by Langley *et al.* [14] to be competitive with Quinlan's popular C4.5 decision tree classifier [20]. Naïve Bayes is so called because it makes the two following, often unrealistic, assumptions:
1. All other variables are conditionally independent of each other given the classification variable
2. All other variables are directly dependent on the classification variable

When represented as a Bayesian network, a Naïve Bayes classifier has a simple structure whereby there is an arc from the classification node to each other node, and there are no arcs between other nodes [8], as in Figure 1(a).

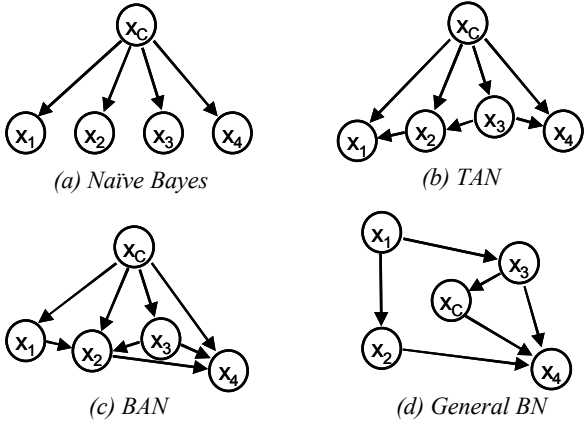

*(a) Naïve Bayes*   *(b) TAN*

*(c) BAN*   *(d) General BN*

**Figure 1:** Illustration of Naïve Bayes, TAN, BAN and General BN Structures

Several researchers have examined ways of achieving better performance than Naïve Bayes by relaxing these assumptions. Friedman *et al.* [8, 9] analyze *Tree Augmented Naïve Bayes* (TAN), which allows arcs between the children of the classification node $x_c$ as in Figure 1(b), thereby relaxing the first assumption above. In their approach, each node has $x_c$ and at most one other node as a parent, so that the nodes excluding $x_c$ form a tree structure. They use a *minimum description length* metric rather than the Bayesian metric used in this paper (though Heckerman *et al.* [11, 12] observe that the two metrics are asymptotically equivalent as the sample size increases). To find arcs between the nodes, they use an algorithm first proposed by Chow and Liu [6] for learning tree-structured Bayesian networks.

Langley and Sage [15] consider an alternative approach called *Selective Naïve Bayes* (SNB), in which a subset of attributes is used to construct a Naïve Bayes classifier. By doing this, they relax the second of the two assumptions listed above. Kohavi and John [13] improve on this by using a wrapper approach to searching for a subset of features over which the performance of Naïve Bayes is optimised.

Cheng and Greiner [4, 5] evaluate the performance of two other network structures. The first is *Bayesian Network Augmented Naïve Bayes* (BAN), in which all other nodes are direct children of the classification node, but a complete Bayesian network is constructed between the child nodes as in Figure 1(c), rather than just a tree. The second is the *General Bayesian Network* (GBN), in which a full-fledged Bayesian network as shown in Figure 1(d) is used for classification. After constructing the Bayesian network, they delete all nodes outside the Markov blanket (see Section 3) prior to using the network for classification. They use an efficient network construction technique based on condition independence tests [3]. They achieve good results with the BAN and GBN algorithms compared with Naïve Bayes and TAN, particularly when a wrapper is used to fine-tune a threshold parameter setting. Baesens *et al.* take a similar approach to Bayesian network classification in a more recent paper: they construct a general Bayesian network using a Markov Chain Monte Carlo approach and then delete all nodes outside the classification node's Markov blanket.

## 2 Induction of Bayesian Networks

In the work described here, the framework developed by Cooper and Herskovits [7] (hereafter referred to as C&H) for induction of Bayesian networks from data is used. This section summarizes their approach to induction of Bayesian network structures and probabilities, and outlines how classification may be performed in this framework.

### 2.1 Determining Network Structure

C&H's framework is built on in determining which of two Bayesian network structures is more likely. If $D$ is a database of cases, $Z$ is the set of variables represented by $D$, and $B_{S_i}$ and $B_{S_j}$ are two belief-network structures containing exactly those variables that are in $Z$, then they aim to calculate $P(B_{S_i}|D)/P(B_{S_j}|D)$. However,

$$\frac{P(B_{S_i}|D)}{P(B_{S_j}|D)} = \frac{\frac{P(B_{S_i},D)}{P(D)}}{\frac{P(B_{S_j},D)}{P(D)}} = \frac{P(B_{S_i},D)}{P(B_{S_j},D)} \quad (1)$$

Therefore, the problem of calculating $P(B_S|D)$ reduces to that of calculating $P(B_S,D)$. C&H's equation for calculating $P(B_S,D)$ is based on four assumptions that they identify:
1. Variables are discrete and all are observed (i.e. there are no hidden or latent variables)
2. Cases occur independently, given a belief network model

3. There are no cases that have variables with missing values
4. The density function $f(B_P|B_S)$ is uniform; we are therefore indifferent regarding the prior probabilities to place on a network structure $B_S$

Let $Z$ be a set of $n$ discrete variables, where a variable $x_i$ in $Z$ has $r_i$ possible value assignments: ($v_{i1}$, ..., $v_{ir_i}$). Let $D$ be a database of $m$ cases, where each case contains a value assignment for each variable in $Z$. Let $B_S$ denote a network structure containing just the variables in $Z$. Each variable $x_i$ in $B_S$ has a set of parents, represented as a list of variables $p_i$. Let $w_{ij}$ denote the $j$th unique instantiation of $p_i$ relative to $D$. Suppose there are $q_i$ such unique instantiations of $p_i$. Let $N_{ijk}$ be defined as the number of cases in $D$ in which variable $x_i$ has the value $v_{ik}$ and $p_i$ is instantiated as $w_{ij}$. Let $N_{ij}$ be defined as:

$$N_{ij} = \sum_{k=1}^{r_i} N_{ijk}$$

Then, given the assumptions outlined above,

$$P(B_S, D) = P(B_S) \prod_{i=1}^{n} \prod_{j=1}^{q_i} \frac{(r_i - 1)!}{(N_{ij} + r_i - 1)!} \prod_{k=1}^{r_i} N_{ijk}! \quad (2)$$

Equation 2 can be combined with Equation 1 to give a computable method of comparing the probabilities of two network structures, when given a database of cases for the variables in the structures. Since, by the third assumption listed above, the prior probabilities of all valid network structures are equal, $P(B_S)$ is a constant. Therefore, to maximize $P(B_S,D)$ just requires finding the set of parents for each node that maximizes the second inner product of Equation 2.

C&H develop this into their K2 algorithm which takes as its input a set of $n$ nodes, an ordering on the nodes, an upper bound $u$ on the number of parents a node may have, and a database $D$ containing $m$ cases. Its output is a list of the parents of each node. The K2 algorithm works by initially assuming that a node has no parents, and then adding incrementally that parent whose addition most increases the probability of the resulting network. Parents are added greedily to a node until the addition of no one parent can increase the network structure probability. The function used in this procedure is taken from the second inner product of Equation 2:

$$g(i, p_i) = \prod_{j=1}^{q_i} \frac{(r_i - 1)!}{(N_{ij} + r_i - 1)!} \prod_{k=1}^{r_i} N_{ijk}! \quad (3)$$

In a single iteration of K2, an arc is added to node $i$ from the node $z$ that maximizes $g(i, p_i \cup \{z\})$. If $g(i, p_i) > g(i, p_i \cup \{z\})$, no arc is added.

## 2.2 Determining Network Probabilities

C&H present a simple formula for calculating conditional probabilities, after the network structure has been found. Let $q_{ijk}$ denote the conditional probability that a variable $x_i$ in $B_S$ has the value $v_{ik}$, for some $k$ from 1 to $r_i$, given that the parents of $x_i$, represented by $p_i$, are instantiated as $w_{ij}$. Then $q_{ijk} = P(x_i=k|p_i=w_{ij})$ is termed a network conditional probability. Let $x$ denote the four assumptions of Section 2.1. Then, given the database $D$, the Bayesian network structure $B_S$ and the assumptions $x$, the expected value of $q_{ijk}$ is given by:

$$E[q_{ijk}|D, B_S, x] = \frac{N_{ijk} + 1}{N_{ij} + r_i} \quad (4)$$

## 2.3 Using a Bayesian Network for Classification

As pointed out by Friedman and Goldszmidt [8], inductive learning of general Bayesian networks is unsupervised in the sense that no distinction is made between the classification node and other nodes — the objective is to generate a network that 'best describes' the data. Of course, this does not preclude their use for classification tasks.

A Bayesian network may be used for classification as follows. Firstly, assume that the value of the classification node $x_c$ is unknown and the values of all other nodes are known. Then, for every possible instantiation of $x_c$, calculate the joint probability of that instantiation of all variables in the network given the database $D$. C&H's formula for calculating the joint probability of a particular instantiation of all $n$ variables is:

$$P(x_1 = X_1, ..., x_n = X_n) = \prod_{i=1}^{n} P(x_i = X_i | p_i = P_i) \quad (5)$$

By normalizing the resulting set of joint probabilities of all possible instantiations of $x_c$, an estimate of the relative probability of each is found. The vector of class probabilities may be multiplied by a misclassification cost matrix, if available.

## 3 Construction of MBBC

As was mentioned in the Introduction, construction of a full Bayesian network for the purposes of classification may be computationally inefficient, as the whole structure may not be relevant to classification. Specifically, classification is unaffected by parts of the structure that lie outside the classification node's *Markov blanket*. As described by Pearl [17], the Markov blanket of a node $x$ is the union of $x$'s direct parents, $x$'s direct children and all direct parents of $x$'s direct children. The Markov blanket of $x$ is one of its Markov boundaries, meaning that $x$ is unaffected by nodes outside the Markov blanket.

Our approach seeks to directly construct an approximate Markov blanket around the classification node. the algorithm involves three steps, as illustrated in Figure 2. In the first step, every node $x_i \in Z - \{x_c\}$ is tested relative to $x_c$ to determine whether it should be considered to be a parent or a child of $x_c$, as follows. If $x_i$ is added as a parent of $x_c$, the overall probability of the network will change by a factor $\mathbf{d}_p$ that is calculated as:

$$\mathbf{d}_p = \frac{g(c, \mathbf{p}_c \cup \{i\})}{g(c, \mathbf{p}_c)} \quad (6)$$

Alternatively, if $x_i$ is added as a child of $x_c$, the overall probability of the network will change by a factor $\mathbf{d}_c$ given by:

$$\mathbf{d}_c = \frac{g(i, \mathbf{p}_i \cup \{c\})}{g(i, \mathbf{p}_i)} \quad (7)$$

Accordingly, by testing whether $\mathbf{d}_p > \mathbf{d}_c$, we can add $x_i$ to either the set of $x_c$'s parent nodes $Z_P$ or its child nodes $Z_C$. However, if $max(\mathbf{d}_p, \mathbf{d}_c) < 1$, no arc is added; $x_i$ is added to the set of nodes $Z_N$ that are not directly connected to $x_c$.

At the end of this first step, having performed this calculation for each node in turn, a set of direct parents ($Z_P$) and direct children ($Z_C$) of $x_c$ have been identified, as shown in Figures 2(b) and 2(c). As originally described [16], this procedure was sensitive to the node ordering, since $\mathbf{p}_c$ changes as parent nodes are added. In the current implementation, this sensitivity to node ordering has been removed by initially calculating $max(\mathbf{d}_p, \mathbf{d}_c)$ for each node, assuming it has no parents, and then sorting nodes in descending order of this quantity. This heuristic seeks to ensure that the nodes that most affect the structure are considered first.

The second and third steps are concerned with completing the Markov blanket by finding the direct parents of $x_c$'s direct children. In the second step, parents are added to the nodes $x_i \in Z_C$ from a set of candidates comprising $Z_P \cup Z_N$, as shown in Figure 2(d). This is done using the K2 algorithm, as was described in Section 2.1. In general, this requires less computational effort than using K2 to construct the entire network structure, as the nodes in this case have been partitioned into mutually exclusive sets of children and candidate parents. Furthermore, the partitioning means that MBBC does not require K2's node ordering.

In the third step, dependencies between the nodes in $Z_C$ are found. Since children of $x_c$ may be parents of other children of $x_c$, such dependencies fall within the Markov blanket of $x_c$. This step is performed by constructing a tree of arcs between the nodes in $Z_C$ as illustrated in Figure 2(e). This is similar to what is done in the TAN algorithm, except that it handles nodes having different sets of parents. Naturally, this step is an approximation, as it can discover at most one additional parent for each node within the group. The procedure followed in the third step is as follows:

(a) A candidate list of arcs is constructed from all permutations of two nodes in $Z_C$
(b) For each pair $(a\ b)$, the quantity $g(a, \mathbf{p}_a \cup \{b\})$ is

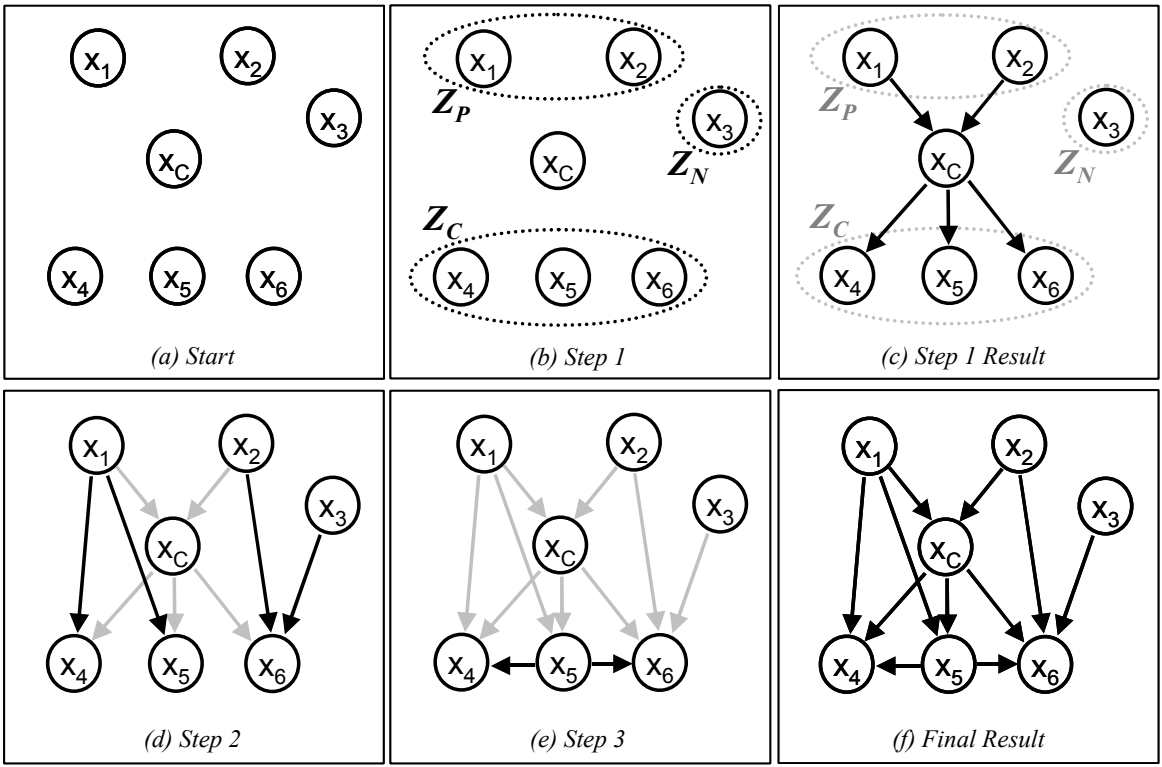

**Figure 2:** Stages in Construction of MBBC

calculated and the pair is deleted from the list if this is less than $g(a, \mathbf{p}_a)$
(c) The list is sorted in order of decreasing $g$
(d) Working sequentially through the list, for each pair ($a$ $b$) an arc is added from $b$ to $a$ unless $a$ is already an ancestor of $b$, and all other pairs beginning with $a$ are removed from the list.

Having constructed the MBBC structure, conditional probabilities are found using the approach of C&H already described in Section 2.2, and the network can then be used for classification as was described in Section 2.3.

## 4 Experimental Results

### 4.1 Methodology

This section describes an experimental evaluation of the MBBC algorithm. It is compared with compared with the Naïve Bayes, TAN and GBN algorithms, all of which were implemented using C&H's inductive learning framework. Accordingly, TAN is implemented as Step 3 of the MBBC algorithm, as described in Section 3. Likewise, the GBN algorithm is actually C&H's K2. Since K2 requires a node ordering, the ordering of variables in the original datasets was used, except that the classification node was placed first so that it could be included as a parent of any other node. For all algorithms (including Naïve Bayes), conditional probabilities were estimated using Equation 4.

The algorithms have been tested on datasets from UCI Machine Learning repository [1]. Since MBBC and K2 require discrete variables and cannot accommodate missing values, datasets were selected that had these characteristics and that were not very small. The datasets are listed in Table 1.

| Dataset | #I | #A |
|---|---|---|
| Chess (King & Rook vs King & Pawn) | 3196 | 32 |
| Wisconsin Breast Cancer Diagnosis | 699 | 9 |
| LED-24 (17 irrelevant attributes) | 3200 | 24 |
| DNA: Splice Junction Gene Sequences | 3190 | 60 |
| Lymphography | 148 | 19 |
| Nursery | 12960 | 8 |
| SPECT Heart Diagnosis | 267 | 23 |
| TicTacToe Endgame | 958 | 9 |

**Table 1:** Datasets Used; Number of Instances (#I) and Number of Attributes (#A) in Each

### 4.2 Accuracy Comparisons

Standard accuracy comparisons were carried out for the four algorithms on all of the datasets. Each dataset was randomly divided into 2/3 for training and 1/3 for testing, and the accuracy of each algorithm on the testing data was measured. Misclassification costs were assumed equal, so that the class predicted in all cases was simply the most probable one. For all except the two datasets with the fewest instances, this procedure was repeated 10 times. For the SPECT and Lymphography datasets, the procedure was repeated 50 times to reduce variability.

Prediction accuracy results and standard deviations are reported in Table 2. Following usual conventions, for each dataset the algorithm with best accuracy is highlighted in boldface. Where two algorithms have statistically indistinguishable performance (based on a paired T-test at the 99% confidence level) and they outperform the other algorithms, they are both highlighted in bold. For example, K2 and MBBC are both best on the DNA Splice dataset and all four are equally good on the Breast Cancer dataset.

|  | **Naïve** | **TAN** | **K2** | **MBBC** |
|---|---|---|---|---|
| **Chess** | 87.63± 1.61 | 91.68± 1.09 | 94.03± 0.87 | **97.03± 0.54** |
| **WBCD** | **97.81± 0.51** | **97.47± 0.68** | **97.17± 1.05** | **97.30± 1.01** |
| **LED-24** | 73.28± 0.70 | 73.18± 0.63 | **73.14± 0.73** | **73.14± 0.73** |
| **DNA** | 94.80± 0.44 | 94.75± 0.42 | **96.22± 0.64** | **95.99± 0.42** |
| **Lymph.** | 83.60± 9.82 | **85.47± 9.49** | 81.47± 10.4 | 83.47± 9.45 |
| **Nursery** | 90.48± 0.41 | **94.16± 0.33** | 92.63± 0.67 | **94.16± 0.33** |
| **SPECT** | 71.70± 6.56 | **81.25± 4.78** | 80.19± 4.66 | 80.75± 4.97 |
| **TTT** | 70.69± 1.94 | 75.08± 1.86 | 74.04± 3.51 | **77.37± 4.37** |

**Table 2:** Average ± Standard Deviation of Percentage Accuracy for All Algorithms and Datasets

Looking at the table, it can be seen that MBBC appears competitive with Naïve Bayes, TAN and K2 for classification tasks. It is best (or joint best) in 7 of the 8 datasets. Its performance is a little worse than TAN on the Lymphography dataset.

### 4.3 ROC Analysis

Although accuracy estimation values such as those in Table 1 are very widely used in the Machine Learning community for comparison of classifiers, Provost *et al.* [19] have argued that accuracy estimation is not the most appropriate metric when cost and class distributions are not specified precisely. As an alternative, they propose the technique of Receiver Operating Characteristic (ROC) analysis, which is taken from signal detection theory. In the Machine Learning context, a ROC graph is a plot of false positives against true positives. A deterministic classifier produces a single point in ROC space, but a probabilistic classifier such as those considered in this paper produces a curve, as the threshold probability over which a positive classification is accepted is varied from 100% to 0%. As stated by Provost and Fawcett [18], the benefit of ROC curves is that they illustrate the behaviour of a classifier without regard to class distribution or error cost.

Figure 3 shows ROC curves for the experiments. There is one graph for each of the eight datasets, and each graph has four curves – one for each classifier.

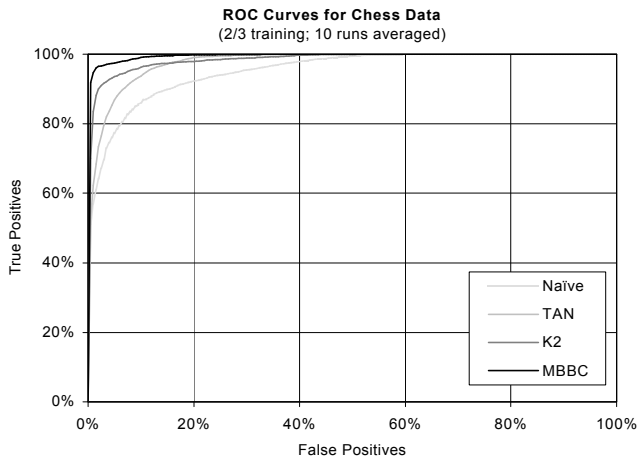
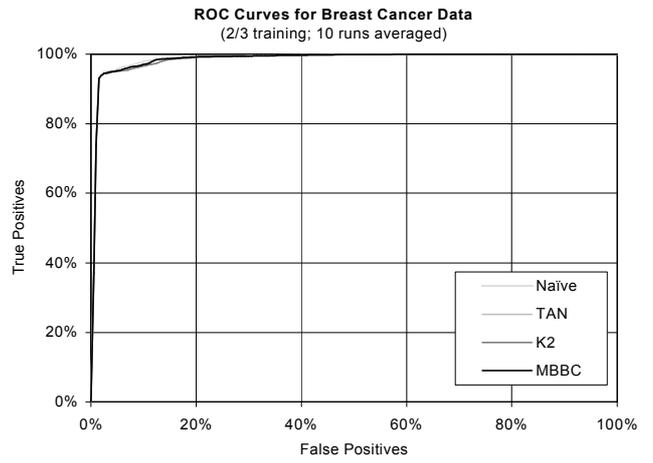
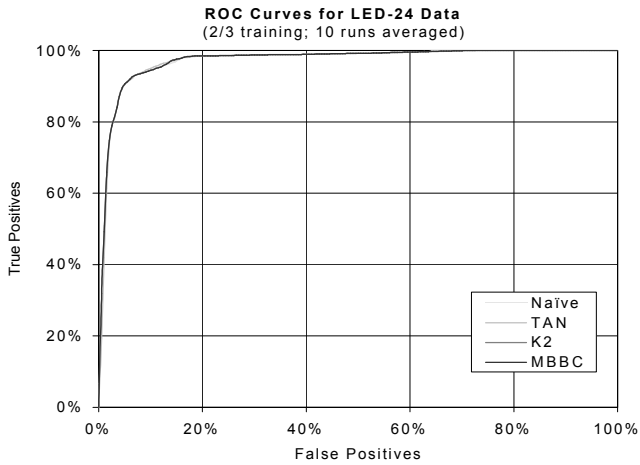
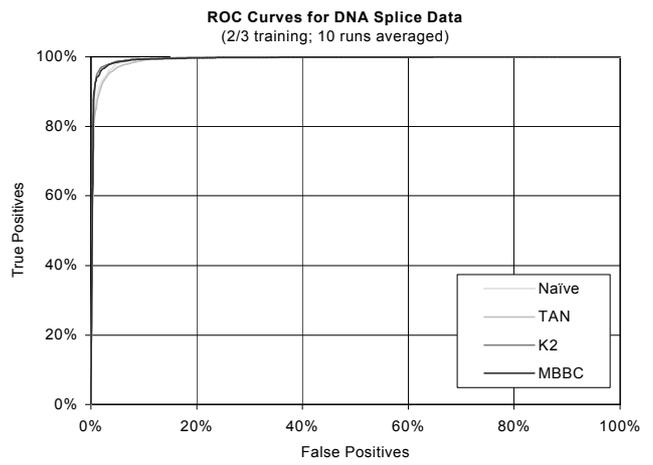
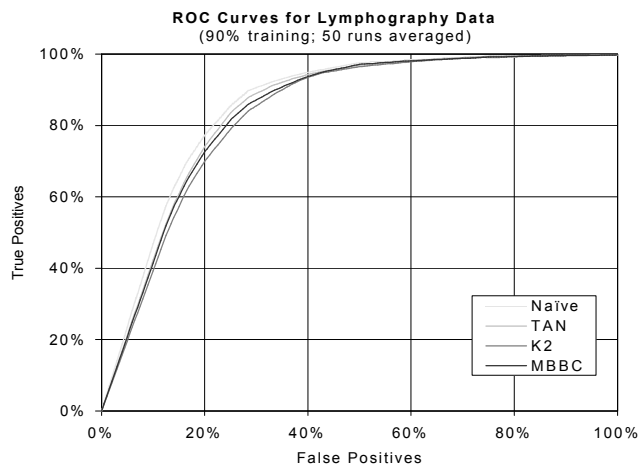
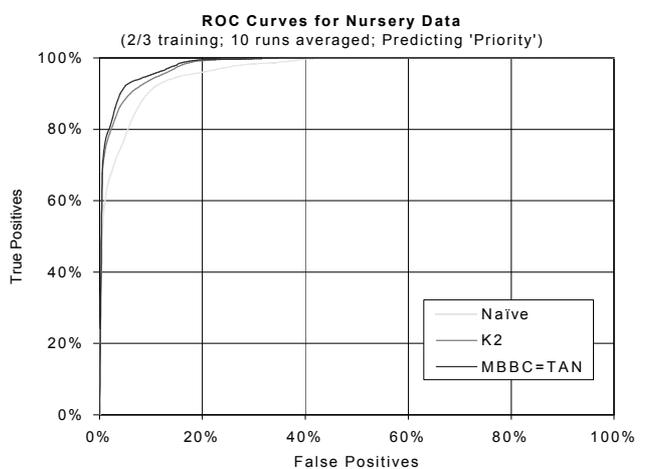
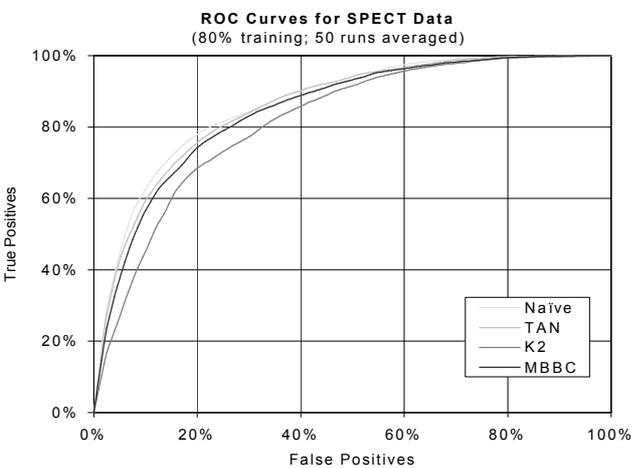
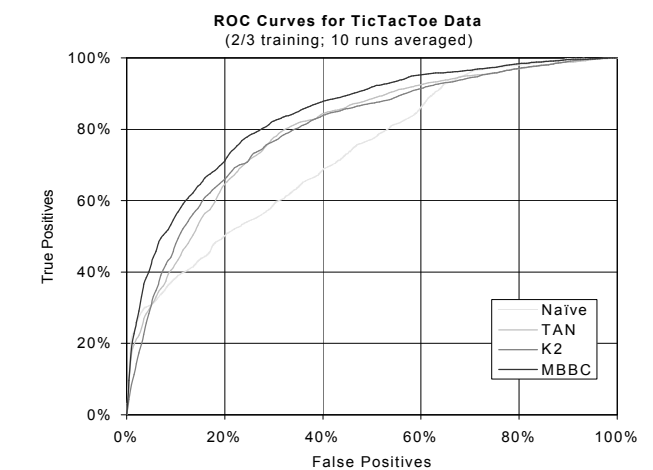

**Figure 3:** ROC Curves for All Datasets

Each individual curve is an average of 10 curves (50 in the cases of the SPECT and Lymphography datasets), one for each of the analyses described above in Section 4.2.

ROC graphs are best suited to two-class problems, which all but one of the datasets are. For the Nursery dataset, the ROC curve is for the prediction of the 'Priority' class.

On a ROC graph, the point (0 0) represents the strategy of never returning a positive classification, no matter how probable, whereas the point (1 1) represents the strategy of always returning a positive classification. The 'ideal' point is the top-left corner at (0 1). In comparing the curves corresponding to two classifiers, one is said to dominate the other if all points on it are to the upper left or equal to corresponding points on the other curve [19].

Looking at Figure 3, there is reasonable correspondence with the results of Table 2. MBBC clearly dominates the other Bayesian classifiers for the Chess and TicTacToe datasets. For the Breast Cancer and LED-24 datasets, there is no real difference in performance between any of the classifiers, based on their ROC curves. For the Nursery dataset, MBBC and TAN jointly dominate the others. Similarly, for the DNA Splice dataset, MBBC and K2 both dominate the others, as reflected in the results of Table 2. For the SPECT and Lymphography datasets, results are more ambiguous and do not correlate well with those of Table 2. In the Lymphography case, Naïve Bayes appears to dominate, although TAN had the highest accuracy in Table 2. The graph for the SPECT dataset is similarly interesting, as it shows that K2 is worse than the other algorithms including Naïve Bayes, even though in Table 2 its accuracy estimation is on par with TAN and MBBC, and better than that of Naïve Bayes.

These two cases illustrate how ROC graphs, by allowing a broader comparison of classifiers than that available from a single-value metric such as accuracy estimation, may reveal different trends in performance.

In general, it is to be expected that MBBC should perform as well as the Naïve Bayes and TAN algorithms, as it is strictly more expressive than these – if the most appropriate representation is a Naïve Bayes or TAN structure, MBBC should be able to find it. It is interesting to note that, on balance, MBBC outperforms K2 even though K2, as a general Bayesian network algorithm, should be as expressive than MBBC. The reason proposed for this are:
(1) K2 requires a node ordering but MBBC does not, so if the given node ordering is not the most appropriate, K2 will discover a sub-optimal structure
(2) Rather than searching for an optimal general BN structure as K2 does, MBBC seeks to optimise the Markov blanket of nodes that affect classification, and thus may be able to discover more subtle dependencies in the data that are specifically relevant to classification.

### 4.4 Speed Comparisons

It is intended that the MBBC algorithm be fast as well as accurate. To evaluate this, its speed in constructing a network has been compared with that of the TAN and K2 algorithms. Naturally, Naïve Bayes could not be included in these comparisons as its structure is fixed.

As a measure of the relative speed of the algorithms, the number of calls to $g$ function (Equation 3) was counted for each algorithm for each dataset. Figure 4 shows the results, with the number of attributes and instances in each dataset listed in brackets. In cases where the number of variables is small, all algorithms are relatively fast and there is not much difference between them. In cases with larger numbers of variables, it can be seen that MBBC's performance scales well, outperforming K2 and sometimes even outperforming TAN.

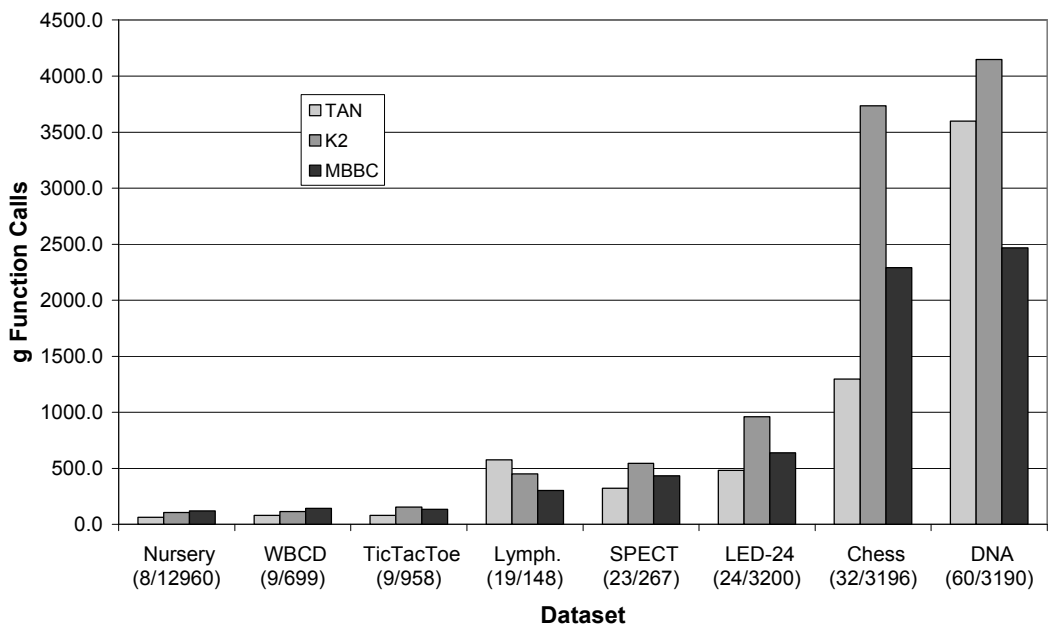

**Figure 4:** Count of Calls to $g$ Function (Equation 3) for TAN, K2 and MBBC

## 5  Conclusions

This paper has presented a description of a Bayesian network structure, called a *Markov Blanket Bayesian Classifier*, for classification tasks, and a method for constructing the MBBC. In this method, an approximate Markov blanket is constructed around the classification node. The method has been implemented using the Bayesian framework for network induction of Cooper and Herskovits [7], although it could be based equally well on any metric for Bayesian networks that has the property of *locality*, whereby networks are scored in terms of their local structure.

A noteworthy feature of Bayesian classifiers in general is their ability to accommodate noisy data: conflicting training examples decrease the likelihood of a hypothesis rather than eliminating it completely.

Key features of the MBBC algorithm are:
- In the first step of constructing the MBBC, all nodes are classified as either parents of the classification node, children of it, or unconnected to it. This contrasts with Naïve Bayes, TAN and BAN structures, where all nodes are children of the classification node. It also contrasts with SNB, where a wrapper-based approach is taken to find which nodes are connected to the classification node.
- In the second and third steps of constructing the MBBC, the only arcs added are to children of the classification node, so that an approximate Markov blanket around the classification node is constructed. This contrasts with GBN structures, in which arcs may be added outside of the Markov blanket but are not considered when using the GBN for classification.
- Unlike K2, the MBBC algorithm does not require an ordering on the nodes.

This paper has also reported on the results of a comprehensive set of experimental comparisons of MBBC with other Bayesian classifiers. These experiments indicate that MBBC is competitive in terms of speed and accuracy with these other classifiers. As was discussed in Section 4.3, MBBC is strictly more expressive than the Naïve Bayes and TAN algorithms, so it is to be expected that it should perform at least as well as well as them. That section also proposed some explanations as to why MBBC can out-perform K2.

In the future, it is hoped to research whether the MBBC approach could be improved by using a different scoring metric such as MDL or conditional independence testing. It is also planned to extend the algorithm to support missing variables (perhaps using the Bound and Collapse approach of Ramoni and Sabastiani [21]), and to support dynamic discretization of continuous variables while constructing the network – the structure of the MBBC should facilitate this, as all variables are associated with the (necessarily discrete) classification node. This could build on previous research by Wu [23] or Friedman and Goldszmidt [9].

All of the algorithms described in this paper have been implemented in Common Lisp and are available for download. For details, please refer to the author's web page: http://www.it.nuigalway.ie/m_madden.

## Acknowledgement

The research described in this paper has been supported by NUI Galway's Millennium Research Programme.

## References

1. Baesens, B., Egmont-Petersen, M., Castelo, R. & Vanthienen, J, 2002: "Learning Bayesian network classifiers for credit scoring using Markov Chain Monte Carlo search," Proc. International Congress on Pattern Recognition.
2. Blake, C.L. & Merz, C.J., 1998. UCI Repository of machine learning databases: http://www.ics.uci.edu/~mlearn/MLRepository.html. University of California at Irvine, Department of Information and Computer Science.
3. Cheng, J., Bell, D.A. & Liu, W., 1997: "Learning Belief Networks from Data: An Information Theory Based Approach." Proc. ACM CIKM '97.
4. Cheng, J. & Greiner, R., 1999: "Comparing Bayesian Network Classifiers." Proc. UAI-99.
5. Cheng, J. & Greiner, R., 2001: "Learning Bayesian Belief Network Classifiers: Algorithms and System." Proc. 14$^{th}$ Canadian Conference on Artificial Intelligence.
6. Chow, C.K. & Liu, C.N., 1968: "Approximating Discrete Probability Distributions with Dependence Trees." IEEE Transactions on Information Theory, Vol. 14, 462-267.
7. Cooper, G.F. & Herskovits, E., 1992: "A Bayesian Method for the Induction of Probabilistic Networks from Data." Machine Learning, Vol. 9, 309-347. Kluwer Academic Publishers, Boston.
8. Friedman, N. & Goldszmidt, M., 1996: "Building Classifiers Using Bayesian Networks." Proc. AAAI-96, Vol. 2, 1277-1284.
9. Friedman, N. & Goldszmidt, M., 1996: "Discretizing Continuous Attributes While Learning Bayesian Networks." Proc. ICML-96.

10. Friedman, N., Geiger, D. & Goldszmidt, M., 1997: "Bayesian Network Classifiers." Machine Learning, Vol. 29, 131-163. Kluwer Academic Publishers, Boston.
11. Heckerman, D, Geiger, D. & Chickering, D.M., 1995: "Learning Bayesian Networks: The Combination of Knowledge and Statistical Data." Technical Report MSR-TR-94-09, Microsoft Corporation, Redmond.
12. Heckerman, D, 1996: "A Tutorial on Learning with Bayesian Networks." Technical Report MSR-TR-95-06, Microsoft Corporation, Redmond.
13. Kohavi, R. & John. G., 1997: "Wrappers for Feature Subset Selection." Artificial Intelligence Journal, Vol. 97, No. 1-3, 273-324.
14. Langley, P., Iba, W. & Thompson, K., 1992: "An Analysis of Bayesian Classifiers." Proc. AAAI-92, 223-228.
15. Langley, P. & Sage. S., 1994: "Induction of Selective Bayesian Classifiers." Proc. UAI-94.
16. Madden, M.G., 2002: "A New Bayesian Network Structure for Classification Tasks." Proc. AICS-02.
17. Pearl, J., 1988: "Probabilistic Reasoning in Intelligent Systems: Networks of Plausible Inference." Morgan Kaufmann, San Francisco.
18. Provost, F. & Fawcett, T., 1997: "Analysis and Visualization of Classifier Performance: Comparison Under Imprecise Class and Cost Distributions." Proc. KDD-97.
19. Provost, F., Fawcett, T. and Kohavi, R., 1998: "The Case Against Accuracy Estimation for Comparing Induction Algorithms." Proc. IMLC-98.
20. Quinlan, J.R., 1993: "C4.5: Programs for Machine Learning." Morgan Kaufmann, San Francisco.
21. Ramoni, M. & Sebastiani, P., 1997: "Learning Bayesian Networks from Incomplete Databases." KMi Technical Report KMi-TR-43.
22. Russell, S. & Norvig, P., 1995: "Artificial Intelligence: A Modern Approach." Prentice-Hall, New Jersey.
23. Wu, X., 1996: "A Bayesian Discretizer for Real-Valued Attributres." The Computer Journal, Vol. 37. No. 8.